# High-precision surgical navigation using speckle structured light-based thoracoabdominal puncture robot

Zezhao Guo [1], Yanzhong Guo [+2], Zhanfang Zhao [*1]

[1] College of Information and Engineering, Hebei GEO University
[2] Beijing Yingrui Pioneer Inc

## Abstract

**Background**

During percutaneous puncture robotic surgical navigation, the needle insertion point is positioned on the patient's chest and abdomen body surface. By locating any point on the soft skin tissue, it is difficult to apply the traditional reflective ball tracking method. The patient's chest and abdomen body surface has fluctuations in breathing and appears irregular. The chest and abdomen are regular and smooth, lacking obvious features, and it is challenging to locate the needle insertion point on the body surface.

**Methods**

This paper designs and experiments a method that is different from previous reflective ball optical markers or magnetic positioning surgical navigation and tracking methods. It is based on a speckle structured light camera to identify the patient's body surface and fit it into a hollow ring with a diameter of 24mm.

**Results**

Experimental results show that this method of the system can be small, flexible, and high-precision positioning of any body surface point at multiple angles, achieving a positioning accuracy of 0.033-0.563mm and an image of 7-30 frames/s.

**Conclusions**

The positioning recognition ring material used in this method can be well imaged under CT, so the optical positioning of the body surface and the in vivo imaging positioning under CT can be combined to form a unified patient's internal and external positioning world coordinates to achieve internal and external registration. Positioning integration. The system senses motion with six degrees of freedom, up and down, front and back, left and right, and all rotations, with sub-millimeter accuracy, and has broad application prospects in future puncture surgeries.

## 1. Introduction

Minimally invasive puncture surgery is a safe and effective solution for the treatment of thoracic and

---

abdominal tumors [1]. Among many medical surgeries, minimally invasive surgery has the characteristics of less trauma, less pain, and faster recovery. Because the current puncture surgery lacks real-time needle path guidance, doctors can only rely on preoperative planning and their own superb skills to perform the surgery is more difficult, and can easily cause complications such as pneumothorax in the lungs. Multiple needle insertions are not allowed. However, doctors generally need to manually puncture about 5 times. More than 8 times means that the operation fails; therefore, robots are introduced in puncture and ablation surgeries. To assist doctors in surgery, it can realize real-time guidance of puncture needles during surgery, greatly improve surgical accuracy and increase the success rate of surgery, which has great research value. Percutaneous puncture surgery robots can reduce complications, improve the accuracy of puncture surgery, and reduce the difficulty of surgery [2].

The thoracoabdominal puncture surgical robot is divided into needle insertion points on the body surface and tumor points in the body according to the surgical path. Tumors in the body generally rely on CT or MRI equipment for observation and positioning, which is relatively mature. However, there is no routine to follow in locating the needle insertion point on the patient's chest and abdomen body surface during puncture surgery navigation. By locating any point on the soft skin tissue, unlike the 3D modeling registration method, the patient's chest and abdomen body surface has changes in the respiratory undulations. It also has an irregular and smooth chest and abdomen, and lacks obvious features. It is difficult to locate the puncture needle point on the body surface. This aspect is rarely discussed in the industry, but this technology is an important part of the surgical navigation of thoracoabdominal puncture robots. This article designed and experimented with a method that is different from previous reflective ball optical markers or magnetic positioning surgical navigation and tracking methods. It is based on a speckle structured light camera to identify the patient's body surface and fit a 24mm hollow ring to locate the patient's chest and abdomen body surface needle entry point. This method can be small, flexible, and high-precision positioning of anybody surface point at multiple angles, including identifying and positioning patients in various sideways positions. Compared with currently used surgical navigation equipment, this method has obvious advantages in dealing with situations where there are ups and downs, and the chest and abdomen are soft, smooth and irregular. The positioning recognition ring material used in this method can be well imaged under CT, so the optical positioning of the body surface and the in vivo imaging positioning under CT can be combined to form a unified patient's internal and external positioning world coordinates to achieve internal and external registration. Positioning integration. The system senses motion in six degrees of freedom, up and down, front and back, left and right, and all rotations, with sub-millimeter accuracy. This real-time positioning of patient displacement can also play a protective role during the treatment process.

Since this field is still in the exploratory stage, this study starts from analyzing the principal characteristics of various current 3D optical positioning technologies, analyzing the technical structures, advantages, disadvantages and applicability of various 3D sensing positioning implementations, and comparing them through accuracy, speed, cost, volume, etc. Indicators were discovered and it was found that the speckle structured light sensor has advantages in this scenario. This study introduces the hardware composition and structure of the speckle structured light camera product; the blind area of the viewing angle during use; the observation rectangle that this method should pay attention to and the blind areas that are easy to produce during observation and the supporting production of software suitable for thoracoabdominal tumor puncture; The results of verifying the accuracy at different distances and different reflective conditions on different GPUs were analyzed, and experiments were conducted on TCP correction and other technologies of this method. In addition, we have done two accuracy verification methods. The first is reprojection error verification, which uses the data collected by the camera to compare with the calibration block to calculate the range of the error. The second is to further combine the operation with the robotic arm and measure the accuracy of precise simulated surgery. Without adding additional workload, the technical accuracy of the re-projection error can reach 0.066-0.13mm, and the speed is 10 frames per second (see the experimental report in the Result section of this article for details). The comprehensive error of the robot arm simulation is in the sub-millimeter range. the following.

This method can establish a clear 3D local model, use the generated point cloud for modeling and registration, and use multiple sensors for multi-angle observation to perceive the operating table environment or the entire surgical scene around the patient to form a multi-degree of freedom full-dimensional observation. With the development of deep learning technology in imaging[3-6]and with the progress of deep learning technology in point clouds[7-10], this technology will have greater breakthroughs in the future.

Collection information obtained by this method is fused with the body surface needle entry point mark coordinates extracted under CT/MRI images to perform internal and external coordinate conversion and matching, and virtually displays the identification of external surgical entry mark points. Positioning selects a path to achieve positioning of tiny areas with a depth of millimeters. These precise marks inside and outside the body form a visual three-dimensional point cloud planning and positioning, and a computer-aided surgery system is constructed. This system integrates the segmentation and positioning coordinate data of internal CT lesion tumors to enable surgical robots to perform percutaneous tumor puncture procedures in the chest and abdomen.

Through the speckle method structured light external real-time imaging of the patient 's marked needle entry point position and posture observation, the patient's breath-hold breathing curve can be obtained, which can be used for Deep inspiration breath hold (DIBH). Because organ motion due to respiration is considered to be the largest intrafraction organ motion, uncertainties during treatment of tumors or lesions affected by respiration must be taken into account. DIBH (maximum inspiration) or shallow BH (moderate inspiration) are effective methods of easing breathing movements. Both concepts minimize tumor motion.

Related work:

Divided into two parts, the history and development of surgical navigation and comparison of current technologies are explained separately.

(1) Historical development of surgical navigation:

In 1908, Horsley and Clark created three-dimensional brain stereotaxic technology and first proposed the concept of navigation. In 1945, Spiegel and Wykes used brain stereotaxic technology to complete the first human brain stereotaxic surgery in history, pioneering the application of navigation in human surgery. In 1986, Roberts in the United States developed the first surgical navigation system. He combined CT images with a surgical microscope and used ultrasound positioning to guide surgery, which was clinically successful. In 1992, Heilbrun et al. in the United States used the principle of binocular stereo vision to apply an image navigation system with infrared tracking technology to clinical applications. This was the world's first optical surgical navigation system [11]. "Surgical navigation" spans a broad field and may be interpreted differently depending on the clinical challenge. The meaning of intraoperative navigation is most accurately defined by the questions asked: "Where is my (anatomical) target?", "How do I reach my target safely?", "Where am I (anatomically)?" or "Where and how should I place my implant?".

(2) Current technology:

Popular surgical navigation tracking and positioning is a marked tracking system based on near-infrared reflective balls and reflective stickers that is widely used[12-15]. This system is mostly used in the field of orthopedics in the field of surgical navigation [16] and relies on near- infrared. The reflective ball system is a tracking system whose main features are optical tracking measurements in medical environments. Mainly by positioning the reflective ball on the machined surgical machinery, the purpose of positioning the end of the surgical instrument is achieved based on the precise correlation of the machining. However, this system is difficult to compare directly with more complex surface-guided radiotherapy programs [错误!未定义书签。Labeled surgical localization, navigation, and modeling become a promising alternative [17]. by placing a dry reflective ball adhered to a soft pad on the body surface, the body surface sphere and the position of the tumor in the body can be well observed under CT images, but it is difficult to fit the specific point, so it is difficult to locate any body surface point. Only the end of the surgical machine and the location of tumors in the body.

Surface Guided Radiation Therapy (SGRT) is a method that uses optical imaging technology to achieve radiation therapy. By detecting the correlation between surface displacement and internal structure displacement, research progress includes fewer body surface markers. In terms of, less restraint and immobilization, safer collision prediction and more accurate real-time tracking[18] , the optical body surface positioning guidance system can process millions of feature points in a few seconds and connect internal lesions to external surgical procedures. The coordinate relationship of the device is registered with the CT image to provide a more comprehensive navigation method, and the accuracy and effect of deep inhalation pause treatment are significantly improved. This method involves a complex setup and calibration process that requires precise setup and calibration to operate effectively, a process that can be time-consuming and require specialized skills. Optical body surface guidance systems can involve high initial investment and maintenance costs.

Electromagnetic tracking systems (EMTS) determine the position and orientation of a target object by measuring a magnetic field with a known geometry. For this purpose, a small sensor coil is inserted into the surgical tool and used to measure the amplitude of the magnetic field generated by a field generator (FG). Magnetic sensors (MS) are very small in size and independent of line of sight, overcoming the limitations of OTS and thus allowing the use of EMTS in many intervention areas such as neurosurgery [19-20] oncology [21], laparoscopy [22]and jaw surgery. Facial Surgery [23]. The accuracy and resolution of these tracking devices are slightly lower than OTS, but are perfectly acceptable for surgical navigation applications. However, the presence of ferromagnetic materials and electronic devices near the magnetic field generator can cause distortion of the reference magnetic field, which is difficult to determine or characterize analytically, and can produce dynamic and static errors in both position and orientation measurements [24-26].

Depending on the clinical setting, the accuracy with which surgical tools are tracked can vary significantly. Currently, structured light technology has been used in brain surgery and spinal surgery in medicine. For example, 7D Surgical's Envision 3DTM and Huake's precise image navigation system can combine intraoperative images with the patient's preoperative CT or MRI images in a few seconds [27]. By projecting a structured light pattern onto the patient's anatomy, using a visible light camera to capture the pattern and performing phase unpacking, a dense 3D point cloud can be obtained, and then a point cloud registration algorithm is used to match the preoperative medical image with the reconstructed point cloud. However, thoracoabdominal puncture still needs further verification.

2. Materials and methods

The 3D imaging and positioning technology in the field of thoracoabdominal puncture surgical navigation uses traditional reflective ball optical markers. In different arrangements and combinations, there are problems such as difficulty in fitting, easy falling off, and poor registration effect. In order to explore new fields with footprints, Effective positioning means, this article analyzes and compares the current 3D imaging and positioning technologies to facilitate in-depth exploration.

2.1 Comparative analysis and discovery of various 3D technology optical sensor technologies:

3D vision imaging is one of the most important methods for robot information collection, and can be divided into optical and non-optical imaging methods. At present, the most widely used optical method is the optical method. Currently, four optical surface scanning technologies are mainly used: laser scanning system [28], time-of -flight system [29-30], stereo vision system [31] and structured light system [32-33].

2.1.1 Structured light technology:

Structured light technology is known for its high precision and is particularly suitable for indoor and close-range scenes. The technology has been studied for applications in orthopedics and neurosurgery. Structured light technology is mainly divided into two types: multi-stripe grating method and speckle method, and the accuracy can usually reach the sub-millimeter or micron level. Compared with the multi-stripe grating method, the speckle method performs better in terms of frame rate and real-time performance, but the accuracy is slightly insufficient. Representative manufacturers of the multi-stripe grating method include Zivid and Kinect2, while representative manufacturers of the speckle method include IDS and Percipio.

2.1.2 Time of flight method (ToF):

ToF technology performs well in long-distance observations, but the error is large at short distances, making it difficult to meet the needs of millimeter-level applications. This technology is mostly used in outdoor scenes, and representative manufacturers include RealSense, Orbec and Sik.

2.1.3 Laser scanning method:

Common laser scanning techniques include triangulation and spectral confocal methods, with micron-level accuracy. In order to obtain better observation results, the observed object and the camera must maintain uniform relative movement. It is usually used in assembly lines, high-speed trains, and shooting of objects moving at a constant speed.

2.1.4 3D stereo vision method [34] : This method includes the following three branches:

(1) Binocular vision technology: relies on the on-site light source environment, has weak anti-**interference** ability, and is suitable for large scenes and large depth of field applications. With the rise of structured light technology, the application of binocular technology in the field of surgical navigation has gradually decreased.

(2) Infrared reflective binocular navigation: Currently widely used in the field of surgical navigation, near-infrared light is used to wirelessly detect and track navigation marks on surgical instruments with sub-millimeter accuracy.

(3) Light field cameras: Light field cameras are relatively costly and offer unique advantages in capturing depth information. As an emerging technology, its potential and limitations in surgical navigation have yet to be further explored and confirmed.

Summarize:

Different types of 3D cameras have their own advantages and disadvantages, depending on the application scenario and needs. High accuracy often results in a small field of view, high speed and large error. In medical fields where precision is required, structured light cameras may be a better choice. Accurate near-end measurements require large errors at long distances, making them suitable for indoor scenes. The cost of three-dimensional light field cameras is too high. the current technology is also mature; in real-time navigation and obstacle avoidance scenarios, ToF cameras and near-infrared navigation cameras have obvious advantages. ToF cameras are fast and are suitable for real-time depth image acquisition. They have high accuracy in short-distance measurements and are suitable for outdoor scenes. Stereo cameras are suitable for depth perception in specific environments, but they require relative movement of objects to be observed. They are suitable for shooting assembly lines or sports and on high-speed trains, stationary or low-speed objects have large errors due to uneven movement speeds; the near-infrared navigation function is suitable for navigation systems such as drones and robots. It cannot generate point clouds for positioning, so in general, the structured light method is more suitable for positioning, modeling, navigation and real-time high-definition images of surgical scenes.

2.2 Explanation of the technical structure of the speckle structured light series: (industry comparative advantages,

hardware structure, observation angle and easy blind spots, supporting software architecture):

2.2.1 Industry comparative advantages:

In order to better choose among structured light technologies, multiple projections, single projections, based on Fourier transform, based on sinusoidal grating stripe projection, deflection method reasoning, based on DeBryan coding, so search the current structured light manufacturers for performance Category ratio, through which the timing is adapted to the technique for surgical scenarios. To compare various manufacturers, in order to facilitate the accuracy and meet the accuracy, the results are deduced according to the speed ranking as shown in Table 1:

**Table 1. Performance comparison of structured light cameras**

Table 1: The gray shaded part is the camera acquisition speed, compare currently available technologies, it can be seen that the speckle speed is fast when the accuracy is met.

| Manufacturer | Shooting speed (shooting + algorithm point cloud shaping)/fps/s | Accuracy (mm) |
| --- | --- | --- |
| NDI (reflective ball marker, no point cloud) | 20-120 | 0.2-0.25 |
| Kinect-v2 | 30 | 2-4mm |
| Kinect-dk | 30-15 days | 11-17mm |
| Realistic (speckle) | 30-90 | 5-14mm |
| ID S (speckle) | 2-30 | Z:025 XY:0.45 |
| Photoneo (grating + speckle) | 20-2.5 | 0.5-0.03 |
| Cognex (speckle) | 10 | 0.25-0.69 |
| Zivid (raster) | 0.1-10 | 0.025 |
| Picture Yang (speckle) | 1 | Z:.025 XY:1.2 |
| KEYENCE ( Grating) Eyes are out of hand | 0.5-1 | 0.2 |
| Mecamander (raster) | 2.5-3 | 0.1 |
| Image (raster) | 1.0-2.0 | 0.15 |
| Bozhong (raster) | 0.2-1 | Z: 0.097 XY: 0.015 |
| Elson (raster) | 0.65-1 | 0.03 |
| Such as this (raster) | 0.87 | 0.17-0.34 |
| Micro chain (grating) | 0.25 | 0.04-0.07 |

Therefore, it can be seen that the accuracy of many products is not suitable for surgical scenes. The infrared marker ball does not generate point clouds. The multi-stripe grating is still mainly based on 12 multi-projections, so it lacks real-time performance. The object cannot be moved during shooting. Some Fourier Leaf grating and speckle methods are more suitable. Therefore, speckle method technology is both accurate and real-time, and some brand pages have established health divisions to focus on medical treatment and have also launched speckle structured light, which shows the future trend. Therefore, a comprehensive study and analysis of the use of speckle structured light technology in the thoracoabdominal meridian Skin puncture field has more advantages.
2.2.2 Speckle structured light camera hardware structure:

After deciding on the speckle method structured light camera, we did a series of experiments and tests to analyze the performance. The following speckle method camera product structure is shown in Figure 1. The product consists of hardware structure including the core module, which is the speckle projection module, including the LED light source. shaping light path, digital speckle glass and industrial lens, which are composed of optical chips and curved mirrors ; the camera acquisition software consists of half-precision matching algorithm and patch algorithm .

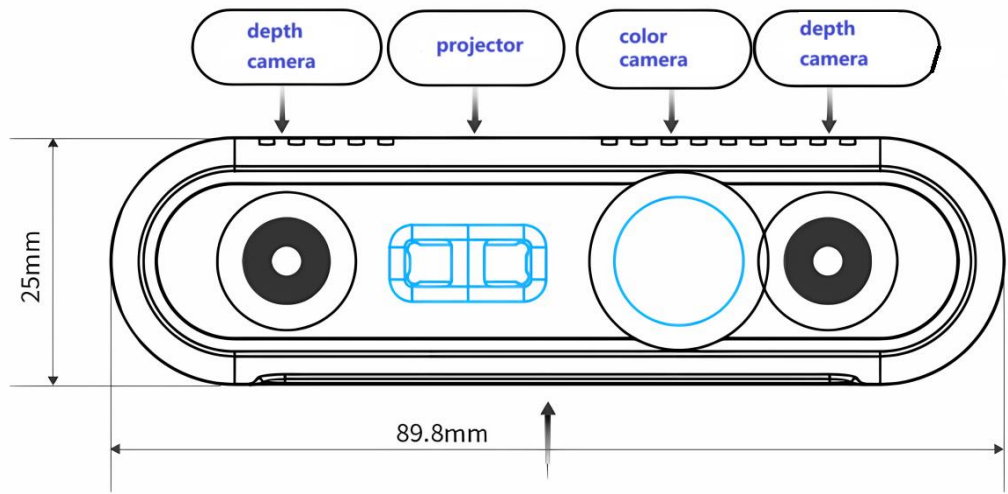

**Figure 1. Speckle camera product components and composition, showing a multi-functional camera module integrating two depth cameras, a central color camera and a projector. The module's compact size is 89.8 mm wide and 25 mm high, and can handle complex image processing capabilities while being compact and easy to integrate.**

2.2.3 Choose the viewing angle and easily create blind spots.

Regarding the observation range of the camera: Because wrong selection of the field of view will make it difficult to improve the accuracy, the observation pyramid consists of the field of view angle and the imaging distance, which is a characteristic of the imaging system. Since a fixed-focus lens is used internally, the zoom also has a zoom range, while maintaining The field of view cannot be changed while maintaining a constant focal length. We discuss two specific locations, "proximal" and "distal." "Near end" is the distance when the imaging distance between the camera and the object being observed is the smallest. The near end position and the final position can be seen in the figure below. The structural optical power is defined by the observation space and the general accuracy range is 1-5% of the observation space. Therefore, choosing the appropriate field of view can achieve ideal accuracy. Use Figures 2 and 3 to illustrate:

Side view height:                    Side view width:

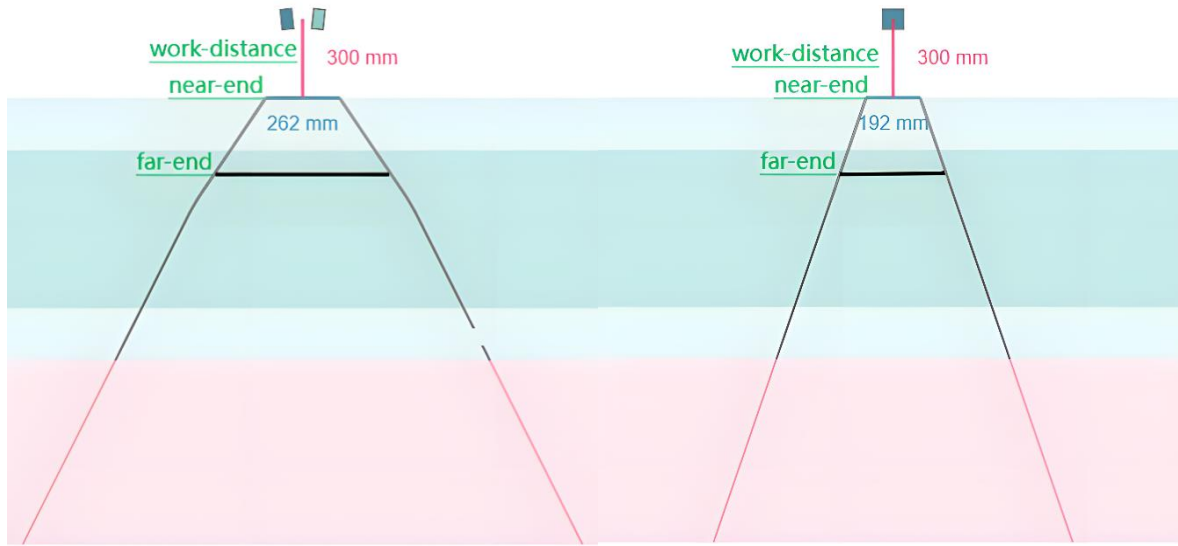

**Figure 2. Suitable surgical observation scene. Analysis of the observation rectangle required for the camera observation range surgical scene.**

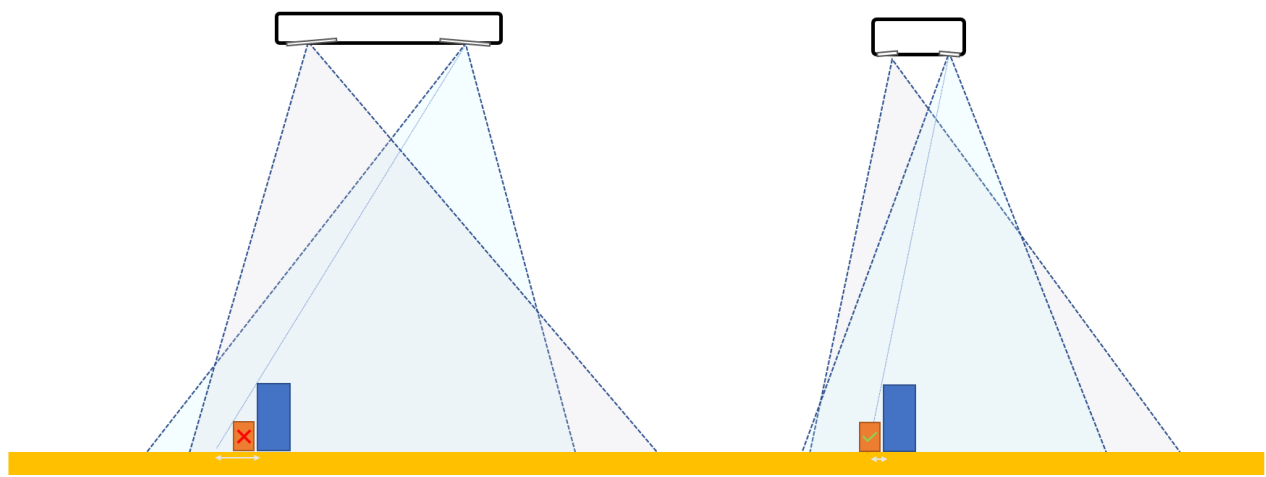

**Figure 3. A schematic diagram that prevents observation due to occlusion , and a comparison schematic diagram of the camera's Field of View (FOV) , showing two different field of view sensors installed at the same height. The sensor on the left has a wide field of view and can detect objects higher up, but cannot detect lower, closer objects. The right sensor has a narrow field of view and can detect low objects close by, but has a smaller range at a distance.**

2.2.4 Software matching:

　　Making camera supporting software is one of the cores of this technology. Through a series of camera calls, hand-eye calibration, and coordinate fusion, a complete set of algorithms and software processes for thoracoabdominal puncture surgery navigation are formed. Therefore, this software process involves camera calls and log records. hand-eye calibration and coordinate fusion and a series of processes. This article uses a flow chart to show the entire thoracoabdominal puncture robotic surgery system process to achieve high-precision robot applications and to accurately calibrate and position the robot's working position (accuracy, speed). The flow chart uses lines of different colors to distinguish Different process modules, and the specific functions or methods inside each module are listed in detail.

Its main components can be divided into the following modules:

Camera calls and logs: Contains related functions for data caching, preventing timeout collection, opening, closing, initialization and reset.

Hand-eye calibration: Lists a series of steps for hand-eye calibration, including setting up the calibration plate, adding and calculating positions, and obtaining coordinate transformations between the camera and hand.

Coordinate fusion: Coordinated fusion related to surgery is mentioned, involving the process of converting positions and calculating robot coordinates.

Identification mark: Introduces how to obtain the mark's position in the robot's reference coordinate system. The software structure process is as shown in Figure 4.

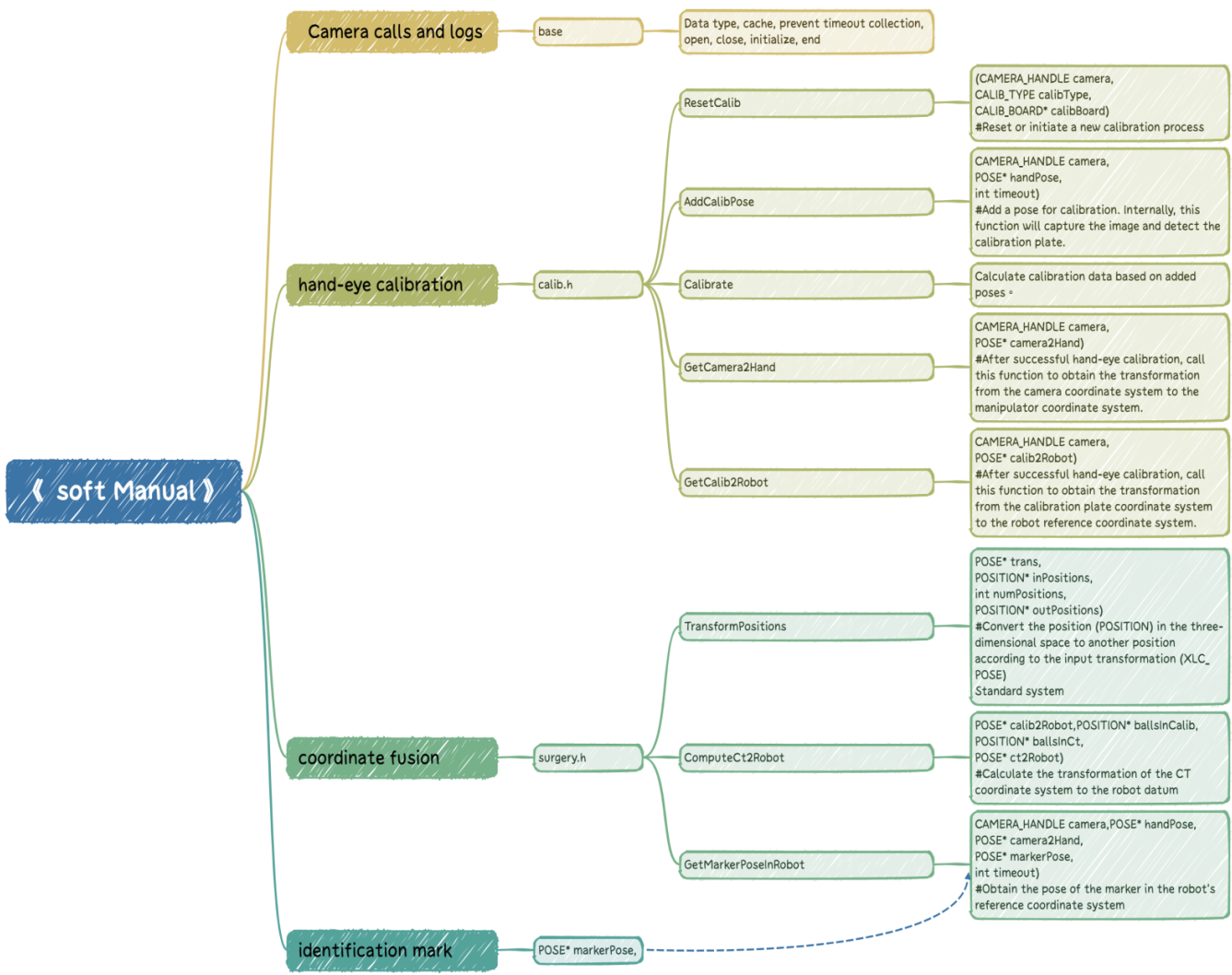

**Figure 4. Split diagram of the functional planning of the surgical navigation system software, showing the software process in the complex robot system, including camera calling, hand-eye calibration, coordinate fusion, and identification mark positioning. The functions listed in each module describe the detailed steps required from initialization to performing precise tasks, indicating the key technical processes in robot navigation and positioning operations.**

3. Result

The speckle method structured light camera is used to achieve an accuracy of 0.0 33 -0. 55 mm and a speed of 3- 30fps /s. Unlike the reflective ball marked optical navigation, which only gives the coordinates of the reflective ball, it also has high-quality 3D images. Various deep learning point cloud 3D image processing can be applied to point cloud information, which can flexibly identify various markers and generate scene semantic information. The algorithm provides high-precision coordinate and posture information of the observed object, provides real-time feedback on scene changes, and is capable of in vitro navigation tasks during surgery. See the experimental results below:

3.1 High-quality imaging shooting effect: The following is the actual shooting effect. The image taken in Figure 5 shows the pleura specially made for the experiment, and a 20mm ring is placed on the chest of the pleura. The imaging effect is shown in the acquisition of 20mm markers without filtering by the algorithm.

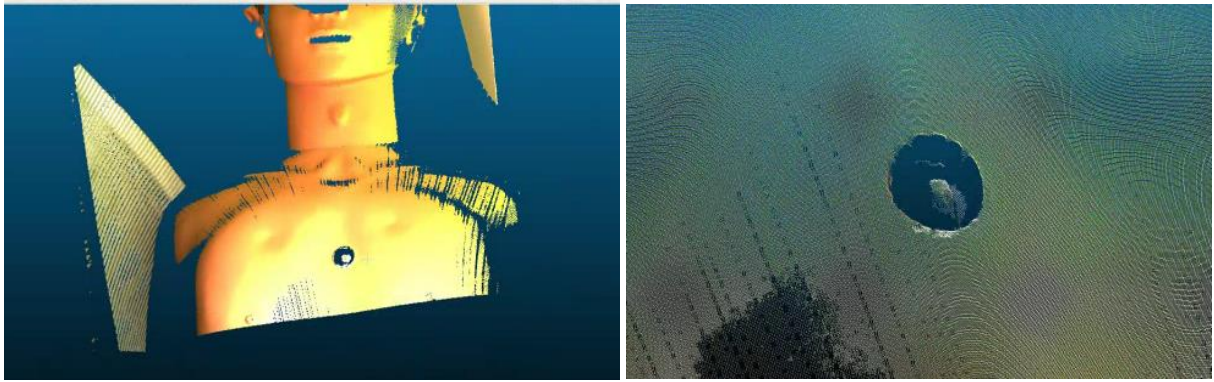

Figure 5. Optical marker image imaging point cloud effect of surgical simulation pleura

3.2 Experimental data results:

The experimental data is used to test and compare the accuracy/optical blur/pixel size and other values in the XYZ direction. The precise measurement effect at different distances is shown in Table 2. It is also explained through comparative data of speed experiments under different graphics cards and different global precision algorithms, see Table 3.

**Table 3**

**Exceeding industry accuracy, the performance parameter table summarizes the vision system's Z-axis accuracy, X- and Y-axis field of view, optical blur (in pixels), and pixels at different viewing distances (in millimeters). Dimensions in millimeters. As the observation distance increases, the accuracy of the Z-axis decreases, the field of view increases, and the optical blur and pixel size change to varying degrees.**

| Observation distance [mm] | Z axis - accuracy [mm] | Field of viewX[mm] | Field of viewY[mm] | Optical blur[pix] | Pixel size [mm] |
|---|---|---|---|---|---|
| 250 | 0.033 | 198.44 | 129.2 | 1.610 | 0.106 |
| 260 | 0.036 | 202.37 | 134.37 | 2.378 | 0.111 |
| 380 | 0.106 | 408.6 0 | 270.68 | 2.377 | 0.223 |
| 400 | 0.117 | 435.37 | 284.93 | 1.937 | 0.234 |
| 500 | 0.183 | 565.23 | 356.16 | 0.262 | 0.293 |
| 6 00 | 0.264 | 658.27 | 427.39 | 1.304 | 0.352 |
| 700 | 0.359 | 751.32 | 498.63 | 2.051 | 0.41 |

| 770 | 0.435 | 816.45 | 548.49 | 2.46 | 0.451 |

### 3.3 Acquisition speed (time of shooting and generating point cloud):

Table 3. Execution speed of global matching and semi-global matching
Note: Will vary with hardware and parameter changes.

| resource | Performance/Quality | CPU Core i7 4x3.6GHz | | GPU Geforce GTx 4 080 | |
|---|---|---|---|---|---|
| | | match | 3D frame | match | 3D frame |
| 1.3MP | SGM 1 (fast) | 230 ms | 280ms | 5.6ms | 11 ms |
| | S GM 8 (accurate) | 460ms | 820 ms | 12.6ms | 62.4 ms |
| | SGM 16 (high resolution) | 190 ms | 750 ms | 21.5 ms | 99 ms |
| 2.3MP | SGM 1 (fast) | 1.0seconds | 1.1 seconds | 120 ms | 230 ms |
| | SGM 8 (accurate) | 2.0 seconds | 2.9 seconds | 180 ms | 0.5 seconds |
| | SGM 16 (high resolution) | 0.8 seconds | 2.5 seconds | 90ms | 0.6 seconds |
| SGM half-precision global matching | | | | 1.2s bandwidth limit | |

3.4 We tested the relationship between point accuracy and working distance, point accuracy and ambient light. The relationship between the performance environment distance and working, and the relationship between speed and graphics card are shown in Figure 5 and Figure 6:

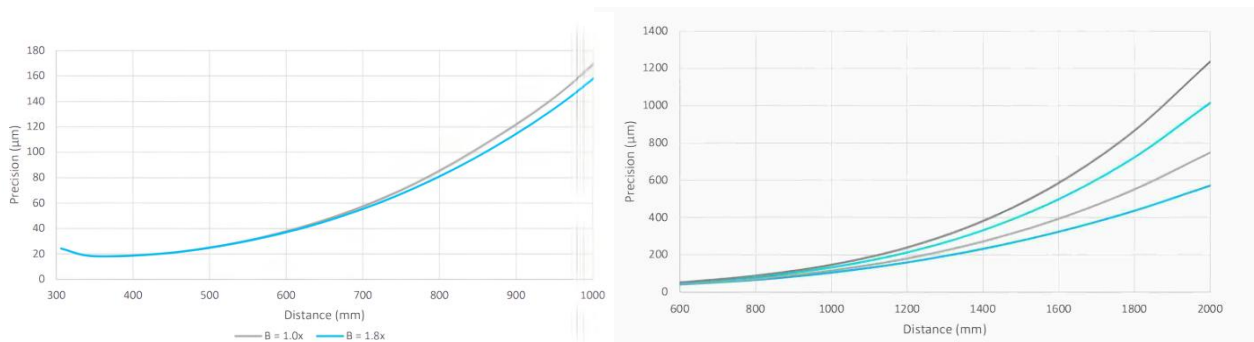

Figure 5. Relationship between point accuracy and ambient light

Figure 6. Relationship between point accuracy and working distance

3.5 Actual accuracy verification (dynamic verification and static verification)

In order to verify the accuracy of this method, we start with two methods: dynamic verification and static verification. Static verification is analyzed through the reprojection error method, which uses the calibration plate image captured by the camera to compare the generated corner values with the standard corner values of the calibration plate to define the resulting error offset value. This method will produce overlapping Pattern and offset standard deviation to verify accuracy, this method is efficient but not intuitive. Dynamic verification uses offline simulation to create the best shooting path, thereby forming a better hand-eye calibration data set, and further forming better results; combined with the TCP tool assembly of the robotic arm, verification is performed according to the execution position of the robotic arm, often by executing multiple Use a needle tip or multiple calibration objects to verify the execution offset value. This method involves many links, so the cumulative error is also large, but it is very intuitive. The execution accuracy of the execution tool can be seen through video and on-site observation. Based on the importance of accuracy in medical care, both methods were adopted in this study. In order to explain the execution process of this method more clearly, and because the final effect needs to be verified in practice, we have made a hand-eye calibration module to make it practical. This module includes software and hardware cooperation verification methods, which are displayed in practice. result. The execution process is as follows:

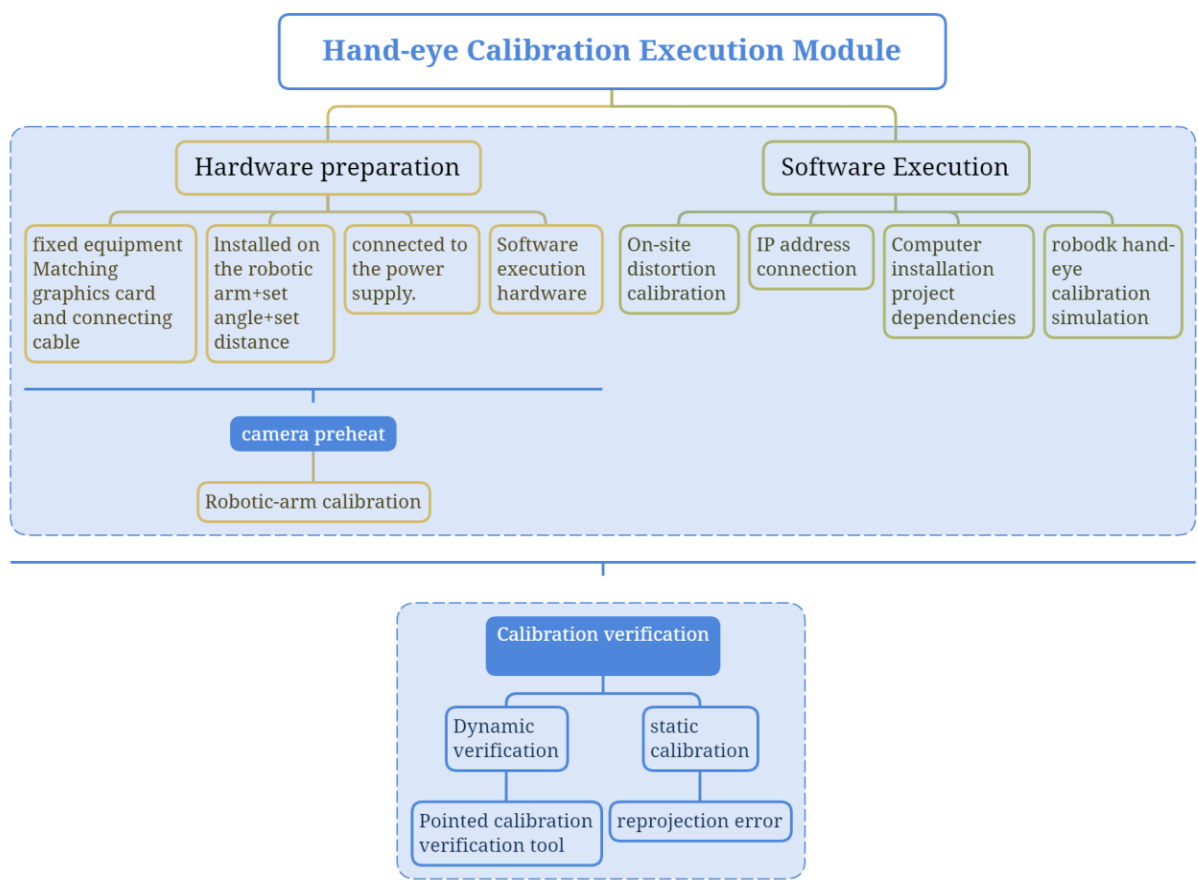

**Figure 7.** The hand-eye calibration module determines the accuracy method and verification method for the final interpretation of accuracy. We designed two accuracy verification methods: reprojection verification calibration and robot arm execution specific pose verification accuracy. In order to implement this goal, we designed both hardware preparation and software execution. The hardware is a camera bracket with horizontal calibration to ensure accuracy. As well as the camera's anti-stress deformation device and higher-precision TCP execution tool, we use 3D printed thick needles. At the software level, RoboDK is used to implement high-quality calibration data sets, so it is necessary to perform simulation of the pose observation rectangle, achieve good calibration poses through simulation modeling, and observe camera gain, ISO, white balance, 3 D image quality, etc. observe high-quality images in each pose to improve the calibration effect. Of course, the residual matrix, rotation error and translation error generated by the calibration are also ablated and compared to obtain better calibration transformation matrix results. Therefore, after obtaining high-quality calibration, the robot arm tcp was used to perform the final point accuracy verification based on the re-projection error, and then the final deviation correction was made based on the effect. The reprojection error is to calculate the deviation value by comparing the corner point information captured by the calibration plate with the real calibration plate data. The robot arm TCP execution is performed using the end of the robot arm tool to the algorithm observation point.

**Since this method includes the robot arm execution error, TCP error, TCP Assembly error, algorithm error, camera observation error, so it is difficult, but it is also the most intuitive. This method directly determines the project execution accuracy**

3.6 Accuracy verification results (linear-based TCP correction):

    The reprojection error will vary every time a third-party module is used. It is always controlled below 0.5 before execution. Since it is not intuitive, this demonstration is performed with the robot arm to execute the final error. It combines multiple error factors., including robotic arm errors, assembly errors, TCP errors, etc., but they are indeed indispensable in reality.

    After the final calibration of the module, the linear correction value of TCP performed by the robot arm reaches the following table 4. The accuracy (linear TCP correction) has a deviation value after TCP point correction. It is concluded that due to the chain derivation forming a linear relationship, linear value correction is carried out to achieve good accuracy performance.

**Table 4 Comparison table of camera observation points and mechanical execution error points Comprehensive final error, robot arm execution point, accuracy after linear correction, the point data is given through camera observation, and then the robot arm executes this point to reach this point, compare the coordinate value given by the camera observation and the robot arm execution to this point the teach pendant displays the point value.**

| Unit (mm) | After moving calibration plate / moving the robotic arm | | | Move fixture and robotic arm again at a large angle | | | The vertical robot arm adjusts position of the calibration plate | | |
|---|---|---|---|---|---|---|---|---|---|
| | Teaching pendant | Starting value | Difference | Teaching pendant | Starting value | Difference | Teaching pendant | Starting value | Difference |
| x | -446.64 | -446.08 | -0.56 | -578.5 | -578.39705 | -0.102943 | -424.45 | -424.53868 | 0.0886814 |
| y | -336.63 | -336.61 | -0.02 | -335.95 | -335.87269 | -0.077305 | -439.78 | -439.67761 | -0.1023870 |
| z | -67.56 | -67.12 | -0.44 | -68.18 | -68.63611 | 0.456114 | -68.38 | -68.764333 | 0.3843331 |
| Total: | | | -1.02 | | | 0.636363 | | | -0.1932646 |
| xyz mean: | | | -0.34 | | | 0.212121 | | | -0.563215 |

### 4. Discussions

    Compared with tracking systems based on optical tracking balls or optical patches, this method has significant advantages in achieving fewer body surface markers, less restraints and fixation, and accurate real-time tracking.

    In the field of thoracoabdominal puncture surgical navigation, soft parts such as the chest and abdomen are large in size and fluctuate with the patient's posture and breathing, resulting in radial positioning changes and large errors in body lesions, making it difficult for traditional tracking systems to fit. This method can more

flexibly adjust the surgical needle insertion point. By using a high-precision, high-real-time speckle structured light system, it can reduce the floating error of lesions in the body caused by observing the respiratory fluctuations based on the body surface. In addition, our solution can process tiny and flexible arbitrary markers through point clouds and 3D images, thereby making surgical navigation and positioning better. This method can also realize real-time observation curves of breathing follow-up, and can be applied to scene observations such as breath-hold observation. Real-time motion monitoring: It can real-time monitor the patient's breathing, coughing, physical discomfort and intra-session motion caused by muscle fatigue during the process, and timely interrupt to avoid treatment deviations.

Generate high-quality point cloud multi-modal images, and use point cloud or multi-modal image processing algorithms to coordinate with the fit to locate the percutaneous puncture point, effectively positioning and executing the percutaneous needle entry point. In addition, this method can combine multiple features of the body surface for rapid registration, and can also be combined with other research, such as [35] Next-generation surgical scene 6-degree-of-freedom scene exploration active surgical global observation, providing local scene needle insertion at the surgical site better implementation.

However, this method also has some limitations. For example, the observation field may become smaller. If the observation space is set according to 300 * 300 * 300, the observation space is small, which may result in a small execution of the observation pose, and other sensors need to be added for global observation. Another limitation is that real-time performance may reduce accuracy. The accuracy of high-quality images is about 7-8 frames, and the accuracy of low-quality images is 30 frames. This method can only track one point on the human body surface, and the monitoring accuracy of respiratory motion is affected by the placement of the marker block.

5. Conclusions

Based on the development of thoracoabdominal puncture surgical navigation technology, this article proposes a surgical navigation method for positioning body surface marker points using speckle structured light. It aims to realize the next generation of robotic surgical navigation in the field of thoracoabdominal puncture to better assist doctors in surgical positioning. Similar structured light precise positioning surgical positioning technology has been actively confirmed in the fields of orthopedics, radiotherapy and other fields. This article has actively explored the field of thoracoabdominal puncture, verified the excellent performance of this method through data analysis and experiments, and analyzed and compared the 3D imaging industry. Based on the principles and characteristics of various technologies and the advantages and disadvantages of each manufacturer's products, it is concluded that speckle structured light has comparative advantages in the field of surgical positioning and navigation. The hardware structure, observation field points and attention methods were analyzed, and the imaging effect of speckle structured light in a simulated pleural surgery environment was verified in the experiment. The results show that the speckle structured light camera is effective in positioning navigation, three-dimensional modeling, and environment perception. Has good experimental performance. The accuracy was verified statically and dynamically through hand-eye calibration. This method is combined with breath-holding to better lock and reduce the floating error of lesions in the body. We also produced a supporting function call table for the camera software, completed a complete experiment through the hand-eye calibration execution architecture, and obtained meaningful experimental results. Finally, we obtained accurate positioning results by comparing the linear TCP modification method of coordinates given by the observation camera with the execution method of the robotic arm. This method has practical implementation significance. The fusion of this method with MRI/CT images can achieve accurate registration in vivo and in vitro, and is fast. Forming a unified 3D high-precision positioning of the external surgical environment and internal organs and lesion points, the speckle structured light system can be used as an important component of the next generation of thoracoabdominal puncture surgical navigation.

Generally, the effect of navigation-assisted surgery is much better than that of the doctor's own surgery [36]. At present, the error of experienced doctors in puncture surgery is also about 5mm. From the experimental performance data table taken on site and calibrated by hand and eye, it can be concluded that the comprehensive accuracy of the speckle structured light system at sub-millimeter precision is 0. 033 -0. 56 submillimeter, which provides an accuracy basis for fusion of MRI/CT coordinate fusion images, making it an important part of navigating surgical instruments to the needle entry point on the body surface. However, due to the industry's difficulty in segmenting lesions in the body, patients are separated at different times. Accurate segmentation and extraction of changes in lesions is still a challenge, and achieving the ultimate goal of fully automatic puncture by machine still requires our unremitting efforts.

**Declaration part**

1) Ethics approval and consent to participate

All participants were fully informed of the purpose, procedures, potential risks, and benefits of the study before the study began, and gave their written consent to participate in this study. All procedures in this study followed the ethical standards of the Declaration of Helsinki.

To protect participant privacy, all data collected is anonymized and results are used for scientific purposes only. Participants may withdraw from the study at any time without penalty or loss of any kind.

2) Consent for publication

Not applicable.

3) Availability of supporting data

The paper data was obtained from Beijing Yingrui Pioneer Medical Technology Inc. The algorithm has been patented and their data cannot be shared based on commercial confidentiality.

4) Competing interests/Authors' contributions

Declaration of interests:

The authors declare that they have no known competing financial interests or personal relationships that could have appeared to influence the work reported in this paper.

Author contributions:

 Ze Zhao Guo contributed most of the creativity/code/experiments, Yan Zhong Guo had equal contributions, did a lot of solid creativity/code/experiments work, and Zhan fang Zhao reviewed the manuscript as a supervisor.

5) Funding

The research work for this article did not receive any external funding support.

6）Declaration-of-competing-interests:

The authors declare that they have no known competing financial interests or personal relationships that could have appeared to influence the work reported in this paper.